\documentclass[10pt,twocolumn,letterpaper]{article}

\usepackage{cvpr}              %

\definecolor{cvprblue}{rgb}{0.21,0.49,0.74}
\usepackage[pagebackref,breaklinks,colorlinks,allcolors=cvprblue]{hyperref}
\usepackage[accsupp]{axessibility}  %

\def\paperID{14} %
\def\confName{CVPR}
\def\confYear{2026}

\title{Investigating Anisotropy in Visual Grounding under Controlled Counterfactual Perturbations}

\author{
  Gabriele Lombardo \quad Luigi Maiorana \quad Liliana Lo Presti \quad Marco La Cascia \\
  Department of Engineering \\
  University of Palermo \\
  Palermo, Italy \\
  {\tt\small \{gabriele.lombardo08, luigi.maiorana02, liliana.lopresti, marco.lacascia\}@unipa.it}
}

\begin{document}
\maketitle
\begin{abstract}
Visual Grounding benchmarks assume that the object described by a referring expression is always present in the image, and grounding models are therefore rarely evaluated under semantically mismatched captions. In such cases, models frequently exhibit approximation behavior, producing a plausible bounding box that satisfies only part of the expression (\eg, preserving the original object while ignoring modified contextual cues). Because mismatched captions represent realistic edge cases, this behavior compromises reliability and raises concerns from an explainability perspective. Identifying its underlying causes is thus essential for improving model faithfulness and interpretability.
Adopting a mechanistic interpretability viewpoint, this work examines whether embedding anisotropy contributes to counterfactual failures. A similarity-controlled counterfactual caption generation protocol is introduced to systematically perturb object or contextual components within predefined embedding similarity intervals, enabling a fine-grained analysis of grounding behavior as a function of alignment. Experiments on two Transformer-based models with markedly different embedding geometries (BERT-based TransVG and CLIP-based SwimVG) reveal no meaningful correlation between cosine similarity and approximation. These findings suggest that anisotropy alone does not account for counterfactual errors, and that robustness requires investigating finer-grained geometric properties of the embedding space.
\end{abstract}
    
\section{Introduction}

Visual Grounding (VG), also referred to as Referring Expression Comprehension (REC), requires a model to localize an object in an image given a text description. Recent Transformer-based architectures have achieved strong performance on standard benchmarks such as \mbox{RefCOCO}~\cite{Yu_Poirson_Yang_Berg_Berg_2016}, RefCOCO+~\cite{Yu_Poirson_Yang_Berg_Berg_2016}, RefCOCOg~\cite{Mao_Huang_Toshev_Camburu_Yuille_Murphy_2016}, and Flickr30K Entities~\cite{Plummer_Wang_Cervantes_Caicedo_Hockenmaier_Lazebnik_2015}. Modern vision-language models~\cite{Li_Wu_Du_Liu_Nghiem_Shi_2025} typically rely on large-scale pretrained models both for vision~\cite{oquab2024dinov, Beal_Wu_Park_Zhai_Kislyuk_2022} and language~\cite{Devlin_Chang_Lee_Toutanova_2019, Min_Ross_Sulem_Veyseh_Nguyen_Sainz_Agirre_Heintz_Roth_2023} to operate in a shared multimodal embedding space.

\begin{figure*}
    \centering
    \includegraphics[width=0.80\linewidth]{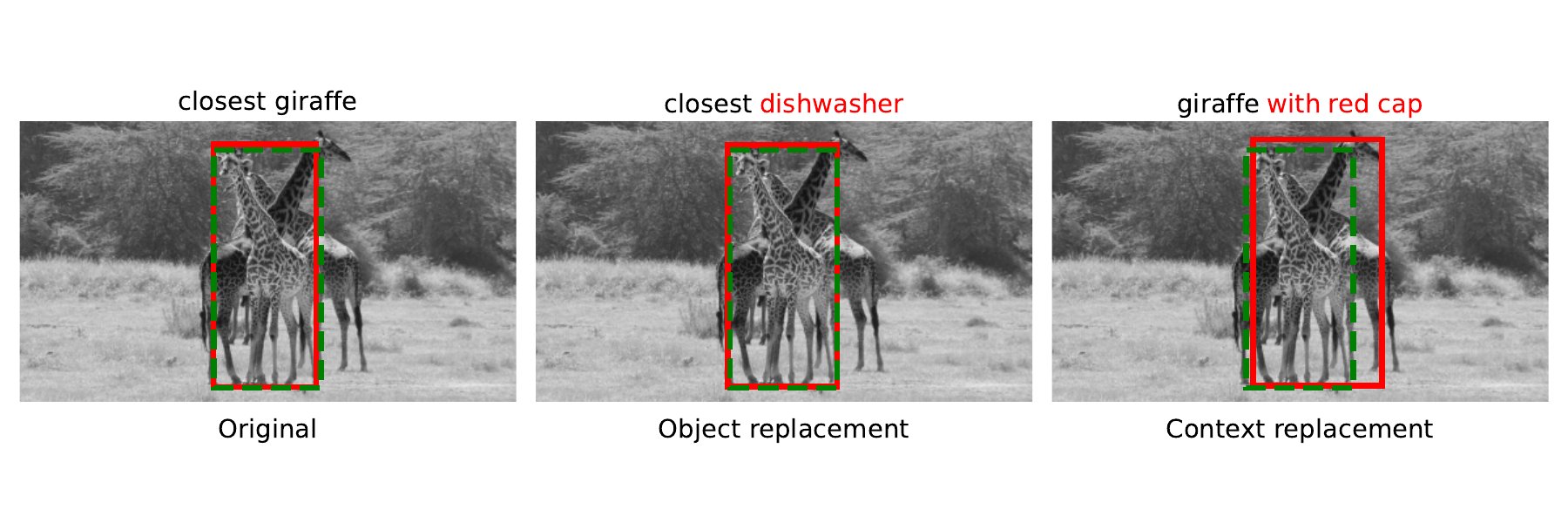}
    \caption{Example of approximation behavior under object replacement and context replacement. In both cases, the model predicts a bounding box that ignores counterfactual cues (highlighted red text) in the modified referring expression. The original RefCOCO ground-truth annotation is shown with a dashed green outline, while the model prediction is drawn as a red bounding box.}
    \label{fig:approx_ex}
\end{figure*}

Despite this progress, current evaluation protocols rely on a strong assumption: the object described in the referring expression is always present in the image. As a consequence, the ability of models to handle counterfactual or mismatched referring expressions is rarely assessed. Furthermore, most grounding architectures are not designed to output calibrated confidence scores or explicit rejection probabilities. When exposed to counterfactual captions, visual grounding models often exhibit \emph{approximation behavior}, \textit{i.e.} predicting a plausible bounding box that satisfies only part of the expression (\eg, grounding the correct object category while ignoring modified contextual cues). An example of this is shown in \Cref{fig:approx_ex}. This approximation behavior does not merely result in localization errors. From an explainability perspective, such predictions undermine the faithfulness of the model's output. In realistic edge-case scenarios, this lack of faithfulness compromises the reliability of the system, particularly when incorrect yet plausible predictions are produced without any indication of uncertainty or rejection.

In this work, we investigate whether the geometric structure of language embedding spaces contributes to approximation in visual grounding models. Prior studies in natural language processing~\cite{Ding_Martinkus_Pascual_Clematide_Wattenhofer_2022, Fuster_Baggetto_Fresno_2022} have shown that contextualized embeddings produced by Transformer architectures exhibit strong anisotropy: representations tend to concentrate along a small number of dominant directions rather than being uniformly distributed in the embedding space. Such geometric biases may reduce discriminability between semantically distinct but lexically similar expressions, potentially affecting a model's ability to detect counterfactual descriptions.

Since current benchmarks do not allow a systematic study of the impact of embedding geometry on grounding behavior, we construct controlled counterfactual datasets derived from RefCOCO+ by systematically replacing either the object or the contextual component of each caption. Importantly, replacements are sampled as controlled perturbations in embedding space to achieve predefined similarity levels with respect to the original caption. This design enables a fine-grained analysis of grounding performance as a function of embedding similarity.

We compare two state-of-the-art Transformer-based grounding models employing language encoders with different geometric properties. We first quantify anisotropy through the average cosine similarity between randomly sampled caption embeddings. We then measure approximation behavior using the Intersection over Union (IoU) between predicted bounding boxes and the original ground-truth annotations under counterfactual conditions.

If anisotropy in language embeddings corresponds to semantically different captions having similar embeddings, one might hypothesize that stronger anisotropy exacerbates approximation.  However, our analysis reveals no clear correlation between embedding geometry and approximation, suggesting that this behavior may emerge from more complex multimodal interactions.

This paper's contributions are:
\begin{itemize}
    \item We introduce a similarity-aware counterfactual caption generation procedure that explicitly accounts for the geometry of the underlying language embedding space.
    \item We propose a quantitative framework to measure the approximation tendency of visual grounding models under controlled counterfactual perturbations.
    \item We provide empirical evidence that embedding anisotropy might not be the main reason for the approximation behavior exhibited by visual grounding models in counterfactual settings.
\end{itemize}

\section{Related Work}
\label{sec:related_work}

\subsection{Visual Grounding}
Visual Grounding~\cite{Deng_Yang_Liu_Chen_Zhou_Zhang_Li_Ouyang_2023, Li_Wang_Yang_Li_Xiao_2025, Shi_Liu_Hu_Hu_Yin_Hong_2025, Ye_Tian_Yan_Yang_Wang_Zhang_He_Lin_2022, Cheng_Liu_He_Ourselin_Tan_Luo_2025, Jin_Luo_Zhou_Sun_Jiang_Shu_Ji_2023} is the vision-language task of localizing a specific object in an image given a natural language description. This task can be seen as the language-grounded extension of object detection, analogous to how Referring Expression Segmentation (RES)~\cite{Ji_Du_Dang_Gao_Zhang_2024} extends object segmentation.

Early visual grounding models~\cite{Yu_Lin_Shen_Yang_Lu_Bansal_Berg_2018, Liu_Wang_Yang_2017} were closely related to conventional object detection architectures and are commonly referred to as two-stage methods.

State-of-the-art visual grounding approaches that are based on the Transformer architecture~\cite{Vaswani_Shazeer_Parmar_Uszkoreit_Jones_Gomez_Kaiser_Polosukhin_2023} are commonly referred to as one-stage methods~\cite{Deng_Yang_Liu_Chen_Zhou_Zhang_Li_Ouyang_2023, Shi_Liu_Hu_Hu_Yin_Hong_2025}. These models are trained end-to-end, and bounding box coordinates are predicted directly from multimodal token representations, typically through dedicated detection/regression heads. While these architectures achieve strong localization performance, they are not explicitly designed to deal with semantic inconsistencies or counterfactual descriptions.

\subsection{Visual Grounding Benchmarks}
Standard evaluation protocols for visual grounding rely on two families of datasets: Flickr30K~\cite{Young_Lai_Hodosh_Hockenmaier_2014} and its extension for visual grounding Flickr30K Entities~\cite{Plummer_Wang_Cervantes_Caicedo_Hockenmaier_Lazebnik_2015}, and \mbox{RefCOCO}~\cite{Yu_Poirson_Yang_Berg_Berg_2016}, RefCOCO+~\cite{Yu_Poirson_Yang_Berg_Berg_2016} and RefCOCOg~\cite{Mao_Huang_Toshev_Camburu_Yuille_Murphy_2016}, all based on the \mbox{MS-COCO}~\cite{Lin_Maire_Belongie_Hays_Perona_Ramanan_Dollar_Zitnick_2014} dataset.

These benchmarks share a strong assumption: the object described in the caption is always present in the corresponding image. Consequently, the ability to reject mismatched captions remains largely unaddressed, preventing a proper assessment of the model's semantically faithful behavior under counterfactual perturbations.

Recent work has attempted to address this limitation by constructing datasets derived from the Ref-COCO family through random caption shuffling~\cite{Kim_Chu_Kurohashi_2022, Li_Wang_Feng_Lyu_Cheng_Li_Liu_Zhao_2023} or by automatically editing captions to describe objects not present in the image~\cite{Fang_Kong_Fowlkes_Yang_2019, Yu_Li_2024}. However, automatically generated counterfactual samples are prone to annotation noise and semantic inconsistencies, which limit their reliability for training and evaluating rejection-aware grounding models. This highlights the need for cleaner benchmarks explicitly designed to evaluate robustness to absent or mismatched referring expressions.

\subsection{Automatic Parsing of Captions}

Recent advances in multimodal dataset construction have been largely driven by the increasing capabilities of vision-language models (VLMs) to bridge visual and linguistic knowledge and generate detailed image descriptions, enabling scalable, automatic collection of annotations for a wide range of multimodal tasks.

In the context of referring expression comprehension and segmentation, the objective is to associate image regions with contextually grounded textual expressions. A common pipeline consists of generating image-level captions, extracting noun phrases via syntactic parsing, and aligning them with visual regions. The authors of~\cite{Xuan_Deng_Ma_2025} first produce both class-agnostic segmentation masks and rich image captions, from which noun phrases are extracted and subsequently matched to regions using CLIP~\cite{Radford_Kim_Hallacy_Ramesh_Goh_Agarwal_Sastry_Askell_Mishkin_Clark_2021}. Similarly,~\cite{Wang_He_Xuan_Sebastian_Ono_Li_Behpour_Doan_Gou_Shen_etal._2024} generate segmentation masks using SAM~\cite{Kirillov_Mintun_Ravi_Mao_Rolland_Gustafson_Xiao_Whitehead_Berg_Lo_etal._2023} with bounding box prompts, where the bounding boxes are obtained by feeding parsed noun phrases into Grounding DINO~\cite{Liu_Zeng_Ren_Li_Zhang_Yang_Jiang_Li_Yang_Su_etal._2025}. 

Related works~\cite{Xiao_Wu_Xu_Dai_Hu_Lu_Zeng_Liu_Yuan_2024, Peng_Wang_Dong_Hao_Huang_Ma_Wei_2023} follow similar strategies to construct bounding box-text pairs, by combining caption generation, phrase extraction, and vision-language alignment.

\subsection{Contextual Embeddings}
Modern vision-language architectures~\cite{Deng_Yang_Liu_Chen_Zhou_Zhang_Li_Ouyang_2023, Shi_Liu_Hu_Hu_Yin_Hong_2025} are Transformer-based and often adopt dual-encoder or cross-encoder architectures, where visual and textual representations are aligned and fused through contrastive objectives or cross-attention mechanisms to enable object localization. However, the geometric structure of these pre-trained embedding spaces, particularly those learned via contrastive objectives such as CLIP~\cite{Radford_Kim_Hallacy_Ramesh_Goh_Agarwal_Sastry_Askell_Mishkin_Clark_2021}, remains a critical factor influencing downstream performance. 

 Isotropy, \textit{i.e.}\@ a uniform distribution of embeddings across all directions of the representation space, has traditionally been regarded as a desirable property of static word embeddings, although it is not consistently observed in all static embedding techniques~\cite{Mimno_Thompson_2017}. 

More recent work has highlighted that contextualized embedding spaces produced by Transformer architectures exhibit anisotropy~\cite{Ethayarajh_2019, Godey_Clergerie_Sagot_2024}, meaning that representations concentrate along a small number of dominant directions rather than being uniformly spread. This geometric bias has been hypothesized to negatively impact semantic discriminability and downstream task performance. However, empirical evidence has questioned this assumption. In particular,~\cite{Fuster_Baggetto_Fresno_2022} report that anisotropy in BERT~\cite{Devlin_Chang_Lee_Toutanova_2019} embeddings is unlikely to be the primary cause of degraded performance in several downstream tasks. Similarly,~\cite{Ding_Martinkus_Pascual_Clematide_Wattenhofer_2022} show that Transformer embedding spaces exhibit a form of \emph{local} isotropy and do not significantly benefit from post-hoc isotropy calibration after training. 

Although anisotropy has been studied in the field of natural language processing, its implications for visual grounding behavior in counterfactual settings remain largely unexplored. Recent open-vocabulary grounding approaches~\cite{Liu_Zeng_Ren_Li_Zhang_Yang_Jiang_Li_Yang_Su_etal._2025, Wang_He_Xuan_Sebastian_Ono_Li_Behpour_Doan_Gou_Shen_etal._2024, Xuan_Deng_Ma_2025} further highlight the importance of robustness to absent or mismatched queries, reinforcing the relevance of our counterfactual setting.

\section{Methodology}

\subsection{Measuring Anisotropy}
\label{sec:measuring_anisotropy}
Given a visual grounding model, we aim to analyze if the anisotropy of the embedding space learned by its language encoder may influence the model's ability to distinguish counterfactual semantic perturbations.

Anisotropy is the concentration of the embeddings along few dominant directions in a hypercone of the embedding space. The greater the anisotropy, the narrower this cone. 

In natural language processing, directions in embedding space are commonly interpreted as encoding semantic information. Accordingly, we adopt cosine similarity as a measure of proximity between representations, as it captures angular alignment independently of vector magnitude, is geometrically well-defined in high-dimensional spaces, and is computationally efficient to evaluate at scale.

Specifically, we compute the cosine similarity between the embeddings of $N$ uniformly sampled pairs of captions. Following~\cite{Ethayarajh_2019}, anisotropy is quantified as the average cosine similarity across all sampled pairs. An average value close to $1$ indicates that randomly selected captions are highly aligned, reflecting a strong concentration of representations within a narrow cone of the embedding space. Such a geometry may reduce directional discriminability, particularly under small semantic perturbations.

\subsection{Building the Approximation Datasets}
To gain insights into the behavior of visual grounding models under counterfactual samples, we construct new datasets derived from the validation and test sets of RefCOCO+. We select RefCOCO+~\cite{Yu_Poirson_Yang_Berg_Berg_2016} among the RefCOCO variants because it excludes absolute location terms (\eg, “left”, “top”), thereby reducing explicit positional bias in the referring expressions.

We denote a sample from RefCOCO+ as a tuple~$\left(\mathbf{I}, t, \mathbf{b}, c\right)$, where $\mathbf{I}$ is the image, $t$ is the caption describing the target object, $\mathbf{b}$ is the bounding box annotation of that object, and $c$ is the corresponding MS-COCO~\cite{Lin_Maire_Belongie_Hays_Perona_Ramanan_Dollar_Zitnick_2014} category. In the context of this work, a caption is composed of two parts: 
(i) the \textit{object}, \textit{i.e.}, the word(s) referring to the category $c$, and
(ii) the \textit{context}, comprising all remaining words that describe the position, attributes, or relations of the referred object.

We build the approximation datasets by systematically replacing either the object or the context of a caption. In contrast to prior approaches~\cite{Yu_Li_2024, Fang_Kong_Fowlkes_Yang_2019}, this replacement process is conditioned on embedding similarity, allowing us to control the degree of perturbation in the language embedding space. These datasets will be used to assess the model's tendency to approximate its predictions under counterfactual inputs, \textit{i.e.}\@, predicting a bounding box that aligns with either only the original object or the original context component of the input caption. The proposed protocol allows us to isolate the effect of semantic similarity while keeping the visual input fixed.

\paragraph{Dependency Parsing}
RefCOCO+ captions are natural language descriptions collected through an interactive game, and objects belonging to the same category are often referred to using different lexical terms (\eg, the \textit{person} category may appear as “man”, “person”, “boy”, \textit{etc}.).
To extract the \textit{object} component of a caption, we use dependency parsing with spaCy~\cite{ines_montani_2023_10009823} to identify the semantic head corresponding to the target entity, also handling multi-word cases, \eg "teddy bear". The \textit{context} is defined as all remaining words that are not part of the object specification.

Building the replacement datasets, we discard captions for which the parser is not able to find a clear semantic head, or multiple candidates are found.

\paragraph{Sampling Replacements}
Directly sampling vectors from the language encoder’s embedding space would not yield semantically valid captions. Instead, we adopt a discrete sampling strategy defined over structured candidate sets.

We first define two vocabularies $\mathcal{O}$ and $\mathcal{C}$, corresponding to candidate objects and contexts, respectively. In our experiments, $\mathcal{O}$ is built from the list of the original \mbox{MS-COCO} categories expanded with common synsets retrieved from WordNet~\cite{Miller_1992} to increase diversity. The context vocabulary $\mathcal{C}$ consists of all context segments extracted from \mbox{RefCOCO+}.

For a given model, in order to obtain controlled perturbation levels, we partition the empirical distribution of cosine similarities between randomly sampled caption embeddings (as defined in Section~\ref{sec:measuring_anisotropy}) into $K$ intervals (in our experiments, we set $K=5$). To ensure an equal number of samples per interval, edges are determined using the distribution quantiles.

Given a caption~$t$ with object component~$o_t$, we sample $K$ replacements, one for each similarity interval. If no candidate is found for a given interval, the caption is discarded. For each candidate~${o \in \mathcal{O}}$, we compute similarity using one of the two following strategies:
\begin{itemize}
    \item \textit{Word-level similarity}: cosine similarity between the embeddings of $o$ and $o_t$.
    \item \textit{Sentence-level similarity}: cosine similarity between the embedding of the original caption~$t$ and that of the modified caption obtained by replacing $o_t$ with $o$.
\end{itemize}
The different strategies for similarity computation allow us to highlight the sensitivity of the embedding space to different perturbations: lexical or contextual respectively. An example of object replacement is shown in \Cref{fig:approx_ex}.

To sample a context replacement for caption $t$, we evaluate each candidate $c \in \mathcal{C}$ by computing the cosine similarity between the embedding of the original caption~$t$ and that of the caption obtained by substituting its context with $c$. A replacement is then sampled within the desired similarity interval. \Cref{fig:approx_ex} shows an example of a caption generated by context replacement.

In summary, we construct three datasets derived from RefCOCO+:
(i) a word-level object replacement dataset,
(ii) a sentence-level object replacement dataset, and
(iii) a context replacement dataset.
Each dataset isolates a different type of controlled perturbation in the caption, allowing us to analyze how object- and context-level modifications affect grounding behavior by keeping the image fixed.

\subsection{Measuring approximation}
From a faithfulness standpoint, a grounding model should modify its prediction when the caption semantics are altered. Predicting the annotated object under counterfactual conditions suggests an approximation behavior.
To evaluate the tendency of a visual grounding model to ignore components of a counterfactual caption and produce a best-match prediction, we leverage both the original RefCOCO+ bounding box annotations and the full set of ground-truth bounding boxes provided by MS-COCO.

In the object replacement setting, we define approximation behavior as the model predicting the original \mbox{RefCOCO+} bounding box despite the caption being counterfactual. To quantify this effect, we compute the Intersection over Union (IoU) between the bounding box predicted under the modified object component and the original \mbox{RefCOCO+} ground-truth annotation. Since the model has no explicit mechanism to output a “not present” decision, it is forced to return a bounding box for every query. A high IoU in this setting indicates that the model fails to revise its prediction in response to the altered object semantics, thereby preserving alignment with the original caption reference rather than following the modified object in the counterfactual caption, which is not present in the image.

In the context replacement setting, approximation occurs when the model ignores the counterfactual context and bases its prediction solely on the (unchanged) object component of the caption. Since RefCOCO+ samples are constructed to distinguish among multiple entities of the same category, several ground-truth bounding boxes of the same MS-COCO category are present in each image. In this case, any of these bounding boxes constitutes an approximation behavior. Accordingly, we define the metric as the maximum IoU between the model prediction and all MS-COCO bounding boxes associated with the target category in the image. In this setting, selecting any instance of the correct category reflects a failure to incorporate contextual information rather than a localization error.

\section{Results}
\subsection{Experimental Setup}
We conduct our experiments on two state-of-the-art transformer-based models.

TransVG~\cite{Deng_Yang_Liu_Chen_Zhou_Zhang_Li_Ouyang_2023} adopts a dual-encoder structure, with separate encoders for both visual and linguistic inputs. The vision and language features are concatenated and fed into a \textit{Vision-Language Fusion Transformer}. A special \texttt{[REG]} token is appended to the full sequence, and the output state of this token is used to directly predict the four bounding box coordinates using a regression head.

SwimVG~\cite{Shi_Liu_Hu_Hu_Yin_Hong_2025} adopts instead a cross-encoder architecture following the early-fusion paradigm in which textual features are injected into the vision encoder to produce caption-aware visual representations. Multimodal interaction therefore occurs directly within the vision branch, removing the need for a separate fusion transformer. The \texttt{[REG]} token is appended to the input sequence of the vision encoder, and its final representation is used by a regression head to predict the bounding box coordinates.

\subsection{Dataset Qualitative Analysis}

During the counterfactual generation process, we apply two filtering stages: 
\begin{enumerate}
    \item if the parser is not able to find a clear semantic head, the caption is discarded.
    \item if no replacement can be found for a given caption in any of the bins listed in \Cref{tab:bin_edges}, the caption is discarded.
\end{enumerate} 

The impact of the filtering pipeline on dataset size is detailed in \Cref{tab:filtered_count}. Since parsing is deterministic, the number of headless captions is identical across all four cases (969); the first row is therefore only reported for completeness.

\begin{figure}
    \centering
    \includegraphics[width=0.7\linewidth]{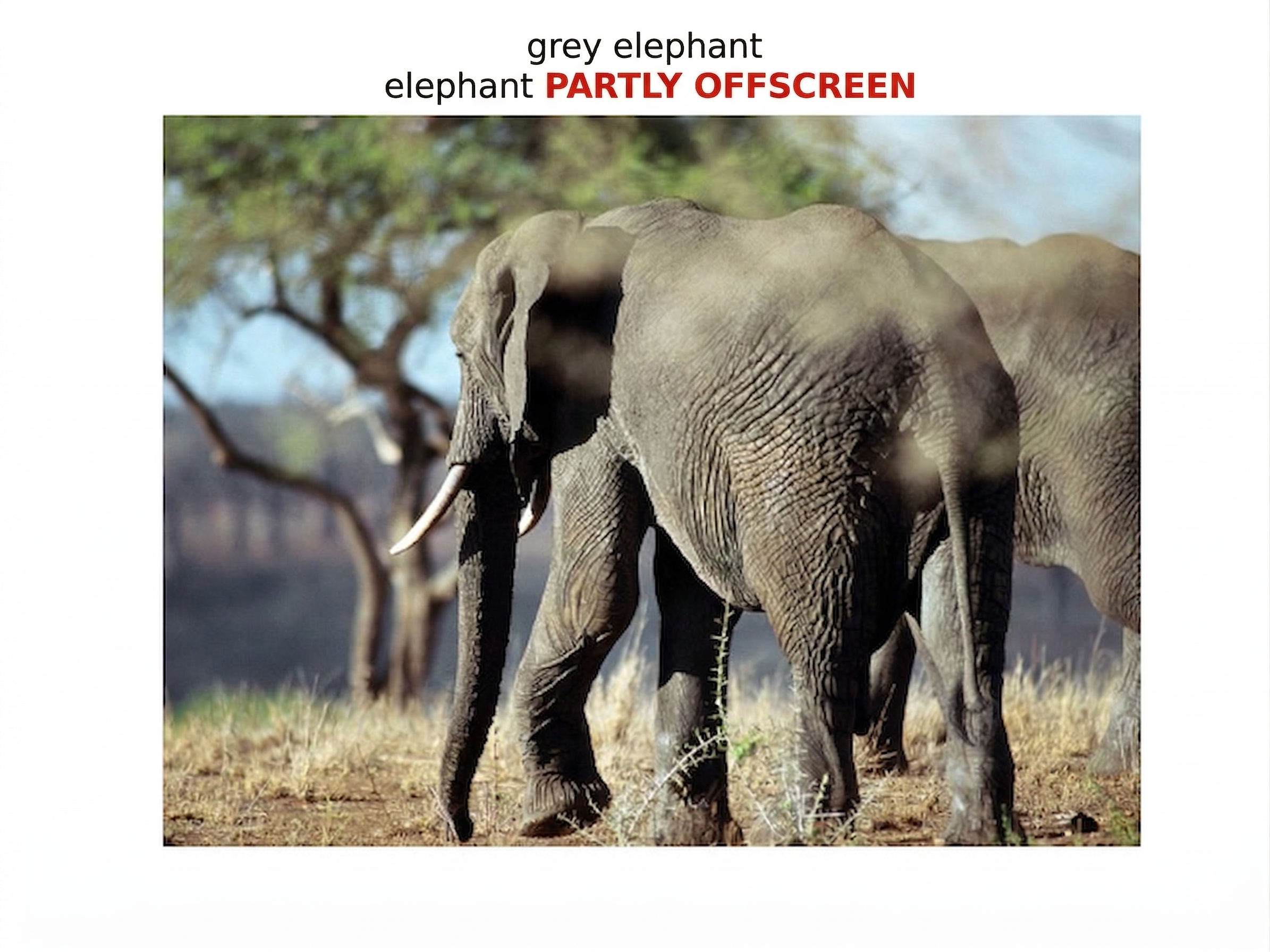}
    \caption{An example of context replacement failure. In this case, the edited caption (bottom row) can still refer to a valid object in the image.}
    \label{fig:ctx_rep_fail}
\end{figure}

As caption generation is fully automatic, concerns may arise regarding the quality of the edited captions. In particular, two potential issues can be identified:
\begin{itemize}
    \item in the \textit{object replacement} setting, the replacement category may still refer to objects present in the image;
    \item in the \textit{context replacement} setting, the modified context may result in a caption that remains valid for the image. An example is shown in \Cref{fig:ctx_rep_fail}.
\end{itemize}
We emphasize that the former case is explicitly avoided in our approach, as replacement candidates are selected by excluding all categories annotated in MS-COCO for the corresponding image.

To evaluate the impact of the latter issue, we conduct a manual validation on a subset of 1,200 samples, uniformly sampled from the two context replacement datasets. This analysis identifies 238 edited captions that remain plausible for the image, indicating that approximately 80\% of the generated samples constitute valid counterfactuals.

\begin{table}
    \centering
    \footnotesize
        \begin{tabular}{c|cccccc}
          Model & 1 & 2 & 3 & 4 & 5 & 6\\ \hline
          SwimVG (CLIP)~\cite{Shi_Liu_Hu_Hu_Yin_Hong_2025} & 0.0 & 0.303 & 0.352 & 0.398 & 0.459 & 1.0 \\
          TransVG (BERT)~\cite{Deng_Yang_Liu_Chen_Zhou_Zhang_Li_Ouyang_2023} & 0.0 & 0.813 & 0.898 & 0.943 & 0.968 & 1.0
        \end{tabular}
    \caption{Edges for the $K=5$ similarity bins computed using the quantiles of the distribution of the cosine similarity of random pairs of captions.}
    \label{tab:bin_edges}
\end{table}

\begin{table}[h]
    \centering
    \setlength{\tabcolsep}{4pt}
    \footnotesize
    \begin{tabular}{l|cc|cc}
         & \multicolumn{2}{c|}{SwimVG} & \multicolumn{2}{c}{TransVG} \\
         & object & context & object & context \\
        \hline
        No semantic head & 969 (13\%) & 969 (13\%) & 969 (13\%) & 969 (13\%) \\
        No replacement & 3604 (55\%) & 795 (12\%) & 6 (0\%) & 2538 (38\%) \\
        \hline
        Caption count & 3005 & 5814 & 4071 & 6603 \\
        Sample Count & 18,030 & 34,884 & 24,426 & 39,618
    \end{tabular}
    \caption{Number of captions discarded after each filtering step. All settings start from 7,578 initial captions. The \textit{Caption count} row reports the number of captions retained after the full filtering pipeline. Each retained caption generates $K + 1$ samples ($K$ counterfactual replacements plus the original), resulting in the final sample counts reported in the last row.}
    \label{tab:filtered_count}
\end{table}

\subsection{Language Encoders Anisotropy}

\begin{figure}[t]
    \centering
    \includegraphics[width=0.8\linewidth]{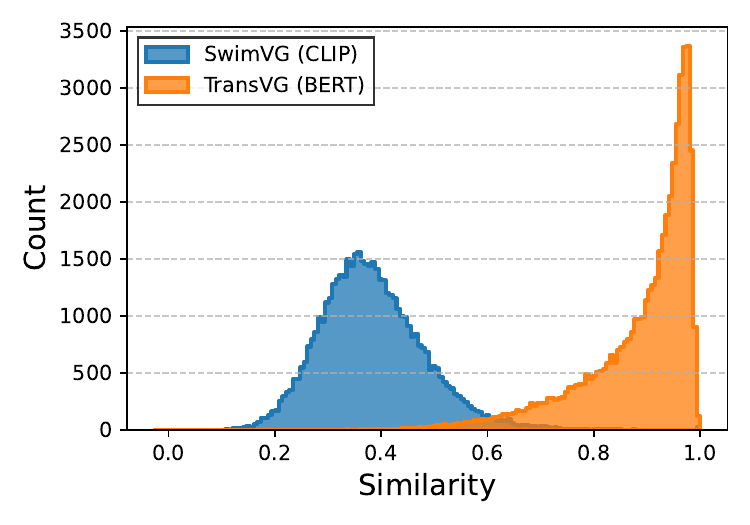}
    \caption{Cosine similarity distribution between the embeddings generated by the language encoders for randomly sampled pairs of captions.}
    \label{fig:anisotropy_comp}
\end{figure}

\Cref{fig:anisotropy_comp} compares the cosine similarity distributions between caption embeddings produced by the two models, computed over $N=50{,}000$ randomly sampled caption pairs from \mbox{RefCOCO+}.

\begin{table}
    \centering
    \begin{tabular}{c|cc|cc}
         Strategy & \multicolumn{2}{|c|}{SwimVG} & \multicolumn{2}{|c}{TransVG} \\
          & $\rho_p$ & $\rho_s$ & $\rho_p$ & $\rho_s$ \\ \hline
         Word-level & $0.055$ & $0.044$ & $0.011$ & $0.011$ \\
         Sentence-level & $0.063$ & $0.038$ & $0.030$ & $0.019$ \\
         \hline\hline
         Context-level & $0.125$ & $0.124$ & $0.007$ & $0.007$ 
    \end{tabular}
    \caption{Correlation coefficients between IoU score and similarity. The first two rows correspond to the object replacement setting, while the last row corresponds to the context replacement setting. $\rho_p$ is the Pearson correlation coefficient, while $\rho_s$ is the Spearman rank-order correlation coefficient.}
    \label{tab:obj_corr}
\end{table}

The average cosine similarity is approximately $0.38$ for SwimVG (CLIP encoder) and $0.88$ for TransVG (BERT encoder). Since perfect isotropy would correspond to an average cosine similarity close to zero, both embedding spaces exhibit anisotropy. However, the substantially higher average similarity observed for BERT indicates a much stronger concentration of embeddings within a narrow cone of the representation space~\cite{Mimno_Thompson_2017}. In contrast, CLIP embeddings are more directionally dispersed due to contrastive multimodal pretraining~\cite{Wolfe_Caliskan_2022}. This behavior is further reflected in the quantiles reported in \Cref{tab:bin_edges}, which show that BERT embeddings are concentrated in the similarity range~$[0.813, 0.968]$. Overall, these results suggest that the BERT language encoder induces a significantly more anisotropic embedding geometry than CLIP. These differences imply that these two models can be used appropriately to assess whether global anisotropy affects a model's approximation behavior.

\subsection{Visual Grounding Approximation}

\begin{figure*}
    \begin{subfigure}[b]{0.40\textwidth}
        \centering
        \includegraphics[width=\textwidth]{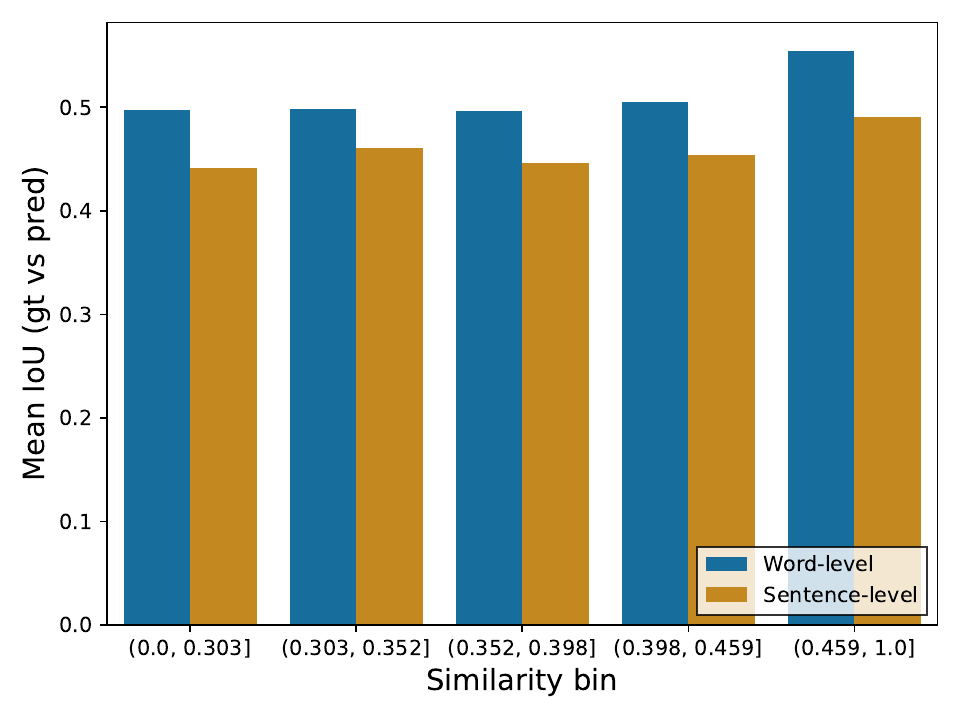}
        \caption{SwimVG}
    \end{subfigure}
    \hfill
    \begin{subfigure}[b]{0.40\textwidth}
        \centering
        \includegraphics[width=\textwidth]{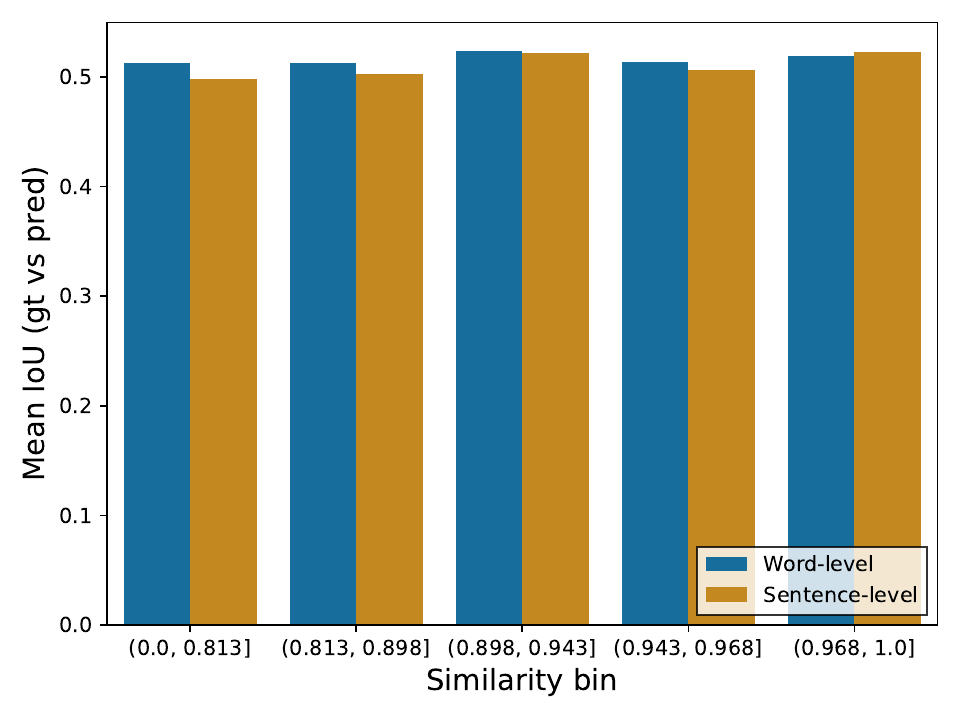}
        \caption{TransVG}
    \end{subfigure}
    
    \caption{Comparison of the mean IoU score per similarity bin in the object replacement case, using both word-level and sentence-level similarity computation. IoU is computed between model predictions and the RefCOCO+ ground-truth bounding boxes.}
    \label{fig:obj_iou_vs_sim}
\end{figure*}

\begin{figure*}
    \begin{subfigure}[b]{0.40\textwidth}
        \centering
        \includegraphics[width=\textwidth]{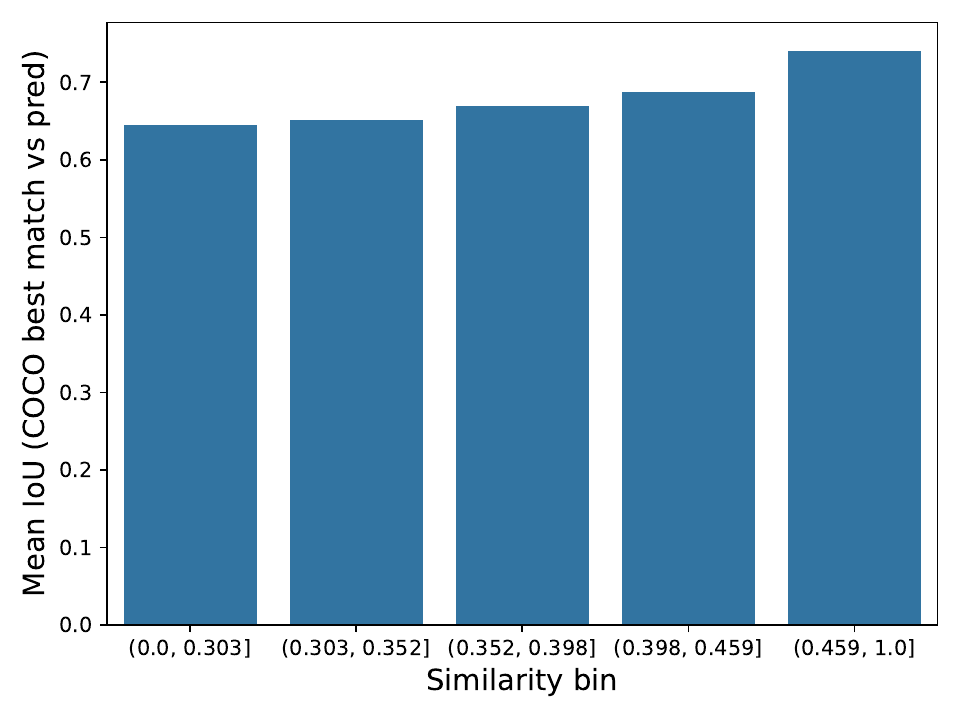}
        \caption{SwimVG}
    \end{subfigure}
    \hfill
    \begin{subfigure}[b]{0.40\textwidth}
        \centering
        \includegraphics[width=\textwidth]{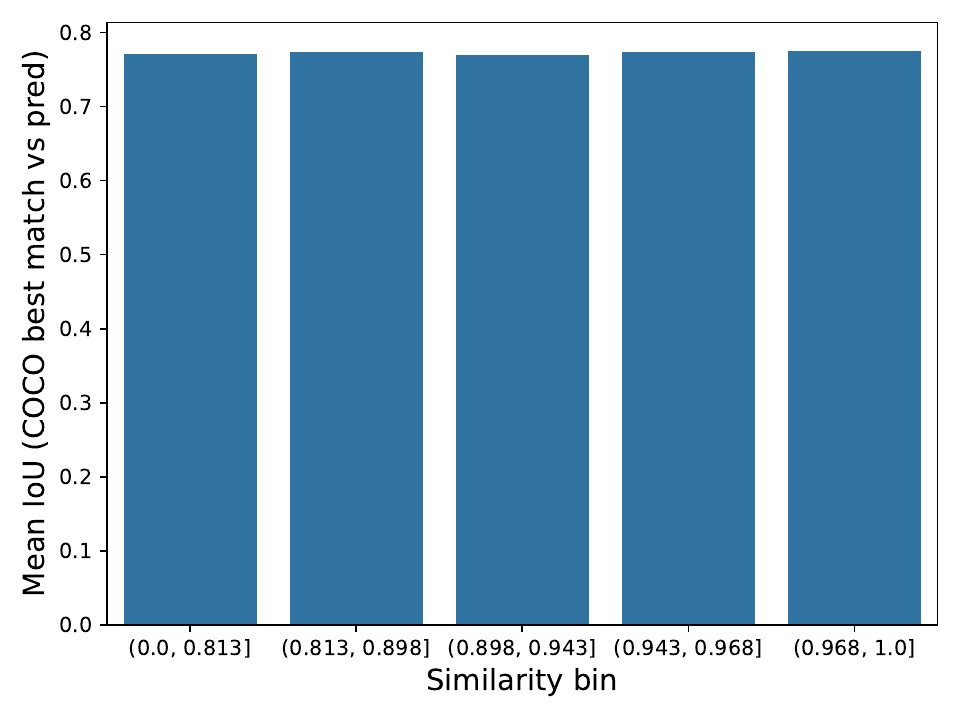}
        \caption{TransVG}
    \end{subfigure}
    
    \caption{Comparison of the mean IoU score per similarity bin in the context replacement case. IoU is computed between model predictions and the best-matching bounding box of the referred category from MS-COCO.}
    \label{fig:ctx_iou_vs_sim}
\end{figure*} 

The results of evaluating the models on the object-replacement and context-replacement datasets are shown in \mbox{\Cref{fig:obj_iou_vs_sim}} and \Cref{fig:ctx_iou_vs_sim}, respectively. Importantly, each similarity bin contains the same number of samples by construction, ensuring a controlled comparison across similarity ranges. In our analysis, we exclude counterfactual captions derived from samples for which the model’s prediction on the original caption achieved an IoU score lower than $0.5$ with respect to the RefCOCO ground-truth bounding box. This filtering ensures that approximation is evaluated only as deviation from the correct grounding behavior.

\begin{figure*}
    \centering
    \includegraphics[width=0.87\linewidth]{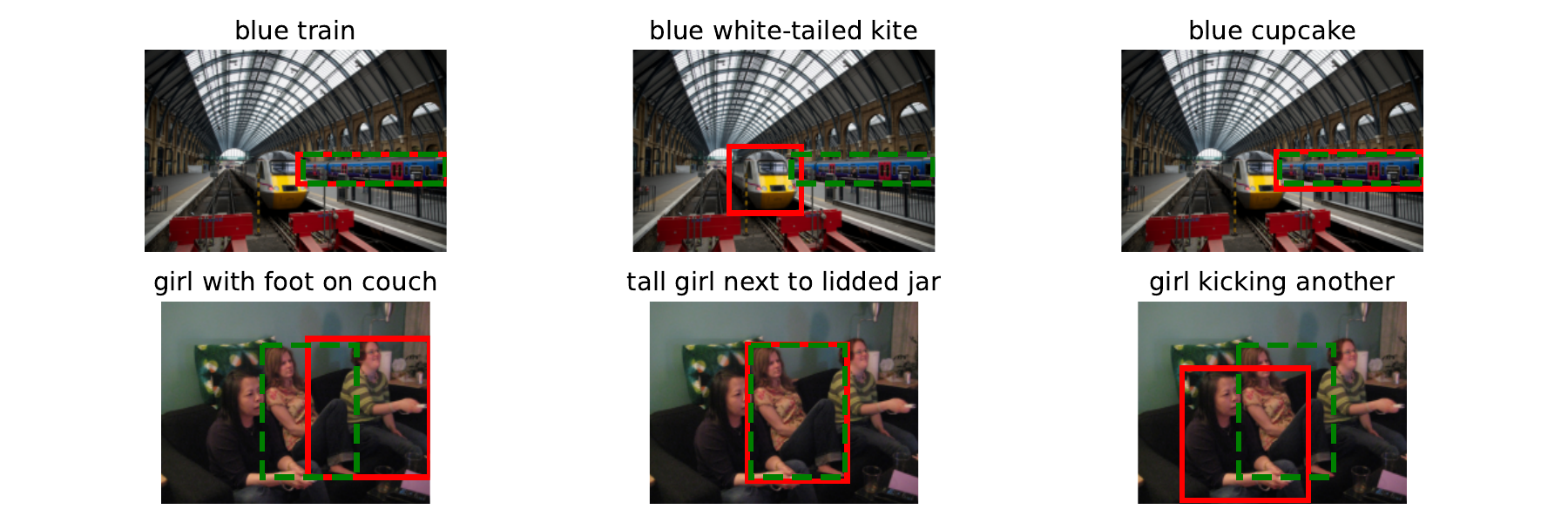}
    \caption{Qualitative results produced by SwimVG on the replacement datasets, showing approximation behavior. The top row corresponds to object replacement, while the bottom row shows context replacement. From left to right, the columns display results for the original caption, the edited caption sampled from the lowest similarity bin, and the edited caption sampled from the highest similarity bin.}
    \label{fig:svg_qualitative}
\end{figure*}

If embedding anisotropy influenced the emergence of approximation behavior, we would expect higher similarity bins showing higher IoU values (as illustrated in the first row of \Cref{fig:svg_qualitative}). Instead, in the object-replacement setting (\Cref{fig:obj_iou_vs_sim}), both TransVG and SwinVG exhibit consistent behavior under word-level and sentence-level similarity computations. As previously stated, TransVG relies on embeddings produced by BERT, whose representations are strongly anisotropic, with embeddings concentrated in a narrow cone of the vector space. In such a geometry, cosine similarity has reduced discriminative power across directions. By contrast, CLIP's representation space used by SwimVG, while not perfectly isotropic, is more directionally dispersed. Despite these geometric differences, neither model shows a meaningful correlation between cosine similarity and localization performance. This observation is quantitatively confirmed by the near-zero Pearson and Spearman coefficients reported in \mbox{\Cref{tab:obj_corr}}. Therefore, while embedding geometry differs substantially across models, anisotropy alone does not explain the visual grounding approximation behavior observed in our experiments. This is consistent with other works in natural language processing~\cite{Fuster_Baggetto_Fresno_2022, Ding_Martinkus_Pascual_Clematide_Wattenhofer_2022} questioning the assumption that isotropy is a crucial property for contextual embeddings.

Comparing \Cref{fig:obj_iou_vs_sim} and \Cref{fig:ctx_iou_vs_sim}, we further observe that localization performance under context replacement is consistently higher than under object replacement across comparable similarity ranges. This suggests that grounding models are more sensitive to perturbations affecting the referential noun than to modifications of contextual information. In other words, altering the semantic anchor of the referring expression leads to stronger degradation than perturbing surrounding descriptive cues, suggesting that grounding decisions are primarily driven by object-level semantics, while contextual cues play a secondary role. From an explainability perspective, this indicates a dominance of noun-driven shortcuts over contextual reasoning.

\begin{figure*}
    \centering
    \includegraphics[width=0.87\linewidth]{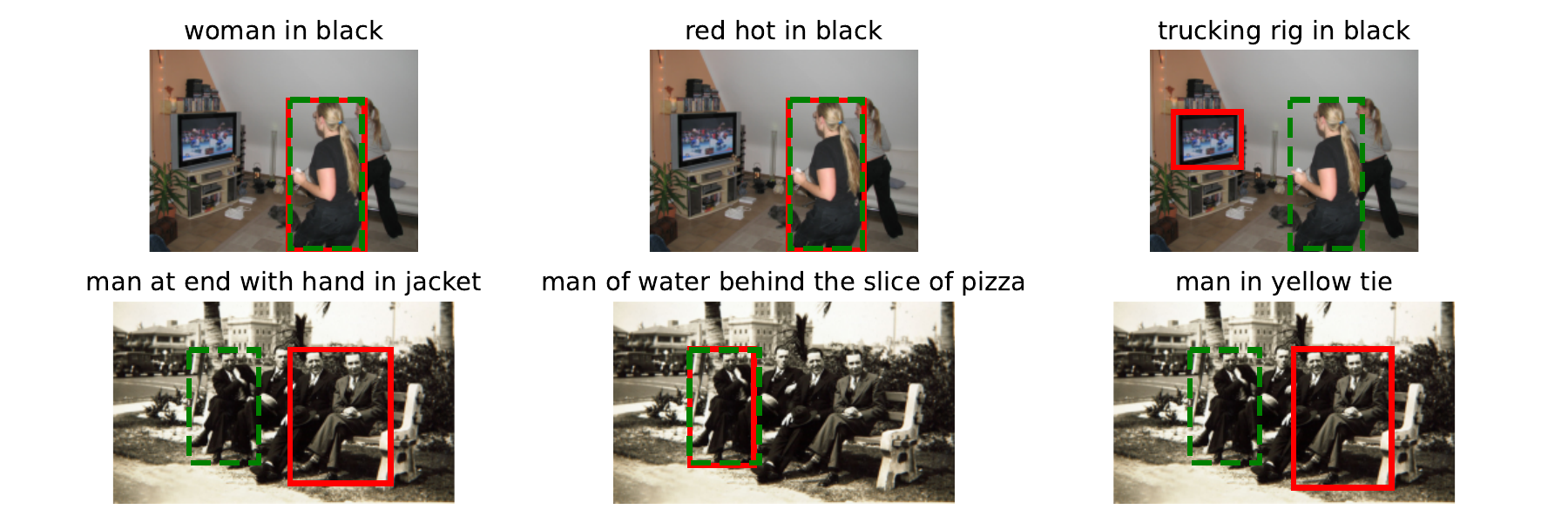}
    \caption{Qualitative results produced by TransVG on the replacement datasets, showing approximation behavior. The top row corresponds to object replacement, while the bottom row shows context replacement. From left to right, the columns display results for the original caption, the edited caption sampled from the lowest similarity bin, and the edited caption sampled from the highest similarity bin.}
    \label{fig:tvg_qualitative}
\end{figure*}

\section{Qualitative results}

\Cref{fig:tvg_qualitative} and \Cref{fig:svg_qualitative} provide qualitative results for both models exhibiting approximation behavior. 

In all but one case (middle top row of \Cref{fig:svg_qualitative}) it is interesting to observe that the model prediction matches the ground-truth annotation under the lowest-similarity perturbation, while it changes under the highest-similarity replacement. Moreover, the middle caption in the second row shows that the model can still rely on the object-level semantic anchor even when the modified caption is not fully coherent from a commonsense perspective.

Moreover, the first row of \Cref{fig:svg_qualitative} exemplifies the behavior that would be expected if anisotropy and approximation were strongly correlated: the model correctly localizes the object for the original caption, fails to reproduce the annotated bounding box under the lowest-similarity replacement, and recovers the same bounding box in the highest-similarity case.

Visual inspection of \Cref{fig:tvg_qualitative} and \Cref{fig:svg_qualitative} indicates that the predicted bounding boxes correspond to existing objects in the image, regardless of the caption semantics. A broader qualitative examination over additional samples reveals that this behavior holds in the majority of cases. This consistency suggests that both models may rely on an internal notion of \mbox{\textit{objectness}}, conceptually analogous to the objectness score in Faster R-CNN~\cite{Ren_He_Girshick_Sun_2016}, which steers predictions toward dominant object regions even under counterfactual inputs.

\section{Conclusions}

In this work, we investigated the role of language embedding geometry in approximation behavior observed in visual grounding models under counterfactual conditions. By introducing a similarity-controlled counterfactual caption generation protocol, we were able to systematically control perturbations in embedding space while keeping the input image fixed and analyze model behavior as a function of semantic proximity by cosine similarity. 

From an explainability standpoint, we highlight that counterfactual robustness is crucial to understanding why visual grounding models fail to revise their predictions when semantically relevant components are altered.

Across models relying on language encoders with different geometric structures of the embedding space, our results reveal only a weak positive correlation between embedding anisotropy and approximation behavior. These findings suggest that such phenomenon cannot be solely attributed to global anisotropy of the embedding space. While our study rules out this factor as a primary cause, further investigation is needed to determine whether grounding failures are better explained by other geometric properties of the language space beyond global anisotropy or instead arise from complex multimodal interactions and the reliance on shortcuts tied to object-level semantic anchors.

Qualitative evidence suggests that both visual grounding models may learn an internal notion of objectness. Nevertheless, confirming this hypothesis requires a systematic and quantitative investigation, which is deferred to future studies.

While our analysis focuses on two representative Transformer-based grounding models with distinct embedding geometries, the observed behavior is likely to extend to other models relying on object-centric representations. Future work should validate these findings on more recent open-vocabulary and cross-encoder models.

Future work may also focus on extending the presented experiments by taking into account other aspects of the embedding space geometry, such as relative distances and cluster density. Changes in the attention dynamics may contribute to the approximation behavior as well, and should be the object of subsequent studies, ultimately aiming to design visual grounding models that are semantically faithful under counterfactual perturbations. 

\section*{Acknowledgement}
This work was partially supported by the European Union under the Interreg NEXT Italie-Tunisie Programme OS 2.4, Project “FORFIRE-AI: Optimisation de la prevention et de la detection des incendies de forest par les techniques de l'Intelligence Artificielle: cas des forets des zones cotier Tunisie-Italie” CUP B73C25000300002.

{
    \small
    \bibliographystyle{ieeenat_fullname}
    \bibliography{main}

\begin{thebibliography}{40}
\providecommand{\natexlab}[1]{#1}
\providecommand{\url}[1]{\texttt{#1}}
\expandafter\ifx\csname urlstyle\endcsname\relax
  \providecommand{\doi}[1]{doi: #1}\else
  \providecommand{\doi}{doi: \begingroup \urlstyle{rm}\Url}\fi

\bibitem[Beal et~al.(2022)Beal, Wu, Park, Zhai, and
  Kislyuk]{Beal_Wu_Park_Zhai_Kislyuk_2022}
Josh Beal, Hao-Yu Wu, Dong~Huk Park, Andrew Zhai, and Dmitry Kislyuk.
\newblock Billion-scale pretraining with vision transformers for multi-task
  visual representations.
\newblock In \emph{2022 IEEE/CVF Winter Conference on Applications of Computer
  Vision (WACV)}, page 1431–1440, Waikoloa, HI, USA, 2022. IEEE.

\bibitem[Cheng et~al.(2025)Cheng, Liu, He, Ourselin, Tan, and
  Luo]{Cheng_Liu_He_Ourselin_Tan_Luo_2025}
Silin Cheng, Yang Liu, Xinwei He, Sebastien Ourselin, Lei Tan, and Gen Luo.
\newblock Weakmcn: Multi-task collaborative network for weakly supervised
  referring expression comprehension and segmentation.
\newblock In \emph{2025 IEEE/CVF Conference on Computer Vision and Pattern
  Recognition (CVPR)}, page 9175–9185, 2025.
\newblock ISSN: 2575-7075.

\bibitem[Deng et~al.(2023)Deng, Yang, Liu, Chen, Zhou, Zhang, Li, and
  Ouyang]{Deng_Yang_Liu_Chen_Zhou_Zhang_Li_Ouyang_2023}
Jiajun Deng, Zhengyuan Yang, Daqing Liu, Tianlang Chen, Wengang Zhou, Yanyong
  Zhang, Houqiang Li, and Wanli Ouyang.
\newblock Transvg++: End-to-end visual grounding with language conditioned
  vision transformer.
\newblock \emph{IEEE Transactions on Pattern Analysis and Machine
  Intelligence}, 45\penalty0 (11):\penalty0 13636–13652, 2023.

\bibitem[Devlin et~al.(2019)Devlin, Chang, Lee, and
  Toutanova]{Devlin_Chang_Lee_Toutanova_2019}
Jacob Devlin, Ming-Wei Chang, Kenton Lee, and Kristina Toutanova.
\newblock Bert: Pre-training of deep bidirectional transformers for language
  understanding.
\newblock In \emph{Proceedings of the 2019 Conference of the North American
  Chapter of the Association for Computational Linguistics: Human Language
  Technologies, Volume 1 (Long and Short Papers)}, page 4171–4186,
  Minneapolis, Minnesota, 2019. Association for Computational Linguistics.

\bibitem[Ding et~al.(2022)Ding, Martinkus, Pascual, Clematide, and
  Wattenhofer]{Ding_Martinkus_Pascual_Clematide_Wattenhofer_2022}
Yue Ding, Karolis Martinkus, Damian Pascual, Simon Clematide, and Roger
  Wattenhofer.
\newblock On isotropy calibration of transformer models.
\newblock In \emph{Proceedings of the Third Workshop on Insights from Negative
  Results in NLP}, page 1–9, Dublin, Ireland, 2022. Association for
  Computational Linguistics.

\bibitem[Ethayarajh(2019)]{Ethayarajh_2019}
Kawin Ethayarajh.
\newblock How contextual are contextualized word representations? comparing the
  geometry of bert, elmo, and gpt-2 embeddings.
\newblock In \emph{Proceedings of the 2019 Conference on Empirical Methods in
  Natural Language Processing and the 9th International Joint Conference on
  Natural Language Processing (EMNLP-IJCNLP)}, page 55–65, Hong Kong, China,
  2019. Association for Computational Linguistics.

\bibitem[Fang et~al.(2019)Fang, Kong, Fowlkes, and
  Yang]{Fang_Kong_Fowlkes_Yang_2019}
Zhiyuan Fang, Shu Kong, Charless Fowlkes, and Yezhou Yang.
\newblock Modularized textual grounding for counterfactual resilience.
\newblock In \emph{2019 IEEE/CVF Conference on Computer Vision and Pattern
  Recognition (CVPR)}, page 6371–6381, 2019.
\newblock ISSN: 2575-7075.

\bibitem[Fuster~Baggetto and Fresno(2022)]{Fuster_Baggetto_Fresno_2022}
Alejandro Fuster~Baggetto and Victor Fresno.
\newblock Is anisotropy really the cause of bert embeddings not being semantic?
\newblock In \emph{Findings of the Association for Computational Linguistics:
  EMNLP 2022}, page 4271–4281, Abu Dhabi, United Arab Emirates, 2022.
  Association for Computational Linguistics.

\bibitem[Godey et~al.(2024)Godey, Clergerie, and
  Sagot]{Godey_Clergerie_Sagot_2024}
Nathan Godey, Éric Clergerie, and Benoît Sagot.
\newblock Anisotropy is inherent to self-attention in transformers.
\newblock In \emph{Proceedings of the 18th Conference of the European Chapter
  of the Association for Computational Linguistics (Volume 1: Long Papers)},
  page 35–48, St. Julian’s, Malta, 2024. Association for Computational
  Linguistics.

\bibitem[Ji et~al.(2024)Ji, Du, Dang, Gao, and
  Zhang]{Ji_Du_Dang_Gao_Zhang_2024}
Lixia Ji, Yunlong Du, Yiping Dang, Wenzhao Gao, and Han Zhang.
\newblock A survey of methods for addressing the challenges of referring image
  segmentation.
\newblock \emph{Neurocomputing}, 583:\penalty0 127599, 2024.

\bibitem[Jin et~al.(2023)Jin, Luo, Zhou, Sun, Jiang, Shu, and
  Ji]{Jin_Luo_Zhou_Sun_Jiang_Shu_Ji_2023}
Lei Jin, Gen Luo, Yiyi Zhou, Xiaoshuai Sun, Guannan Jiang, Annan Shu, and
  Rongrong Ji.
\newblock Refclip: A universal teacher for weakly supervised referring
  expression comprehension.
\newblock In \emph{2023 IEEE/CVF Conference on Computer Vision and Pattern
  Recognition (CVPR)}, page 01–10, 2023.
\newblock ISSN: 2575-7075.

\bibitem[Kim et~al.(2022)Kim, Chu, and Kurohashi]{Kim_Chu_Kurohashi_2022}
Yongmin Kim, Chenhui Chu, and Sadao Kurohashi.
\newblock Flexible visual grounding.
\newblock In \emph{Proceedings of the 60th Annual Meeting of the Association
  for Computational Linguistics: Student Research Workshop}, page 285–299,
  Dublin, Ireland, 2022. Association for Computational Linguistics.

\bibitem[Kirillov et~al.(2023)Kirillov, Mintun, Ravi, Mao, Rolland, Gustafson,
  Xiao, Whitehead, Berg, Lo, Dollár, and
  Girshick]{Kirillov_Mintun_Ravi_Mao_Rolland_Gustafson_Xiao_Whitehead_Berg_Lo_etal._2023}
Alexander Kirillov, Eric Mintun, Nikhila Ravi, Hanzi Mao, Chloe Rolland, Laura
  Gustafson, Tete Xiao, Spencer Whitehead, Alexander~C. Berg, Wan-Yen Lo, Piotr
  Dollár, and Ross Girshick.
\newblock Segment anything.
\newblock In \emph{2023 IEEE/CVF International Conference on Computer Vision
  (ICCV)}, page 3992–4003, 2023.
\newblock ISSN: 2380-7504.

\bibitem[Li et~al.(2025{\natexlab{a}})Li, Wang, Yang, Li, and
  Xiao]{Li_Wang_Yang_Li_Xiao_2025}
Hongbing Li, Xinran Wang, Linyi Yang, Qi Li, and Bo Xiao.
\newblock Sala: Semantic alignment and localization alignment for visual
  grounding.
\newblock \emph{Neurocomputing}, 654:\penalty0 131265, 2025{\natexlab{a}}.

\bibitem[Li et~al.(2023)Li, Wang, Feng, Lyu, Cheng, Li, Liu, and
  Zhao]{Li_Wang_Feng_Lyu_Cheng_Li_Liu_Zhao_2023}
Menghao Li, Chunlei Wang, Wenquan Feng, Shuchang Lyu, Guangliang Cheng,
  Xiangtai Li, Binghao Liu, and Qi Zhao.
\newblock Iterative robust visual grounding with masked reference based
  centerpoint supervision.
\newblock In \emph{2023 IEEE/CVF International Conference on Computer Vision
  Workshops (ICCVW)}, page 4653–4658, 2023.
\newblock ISSN: 2473-9944.

\bibitem[Li et~al.(2025{\natexlab{b}})Li, Wu, Du, Liu, Nghiem, and
  Shi]{Li_Wu_Du_Liu_Nghiem_Shi_2025}
Zongxia Li, Xiyang Wu, Hongyang Du, Fuxiao Liu, Huy Nghiem, and Guangyao Shi.
\newblock A survey of state of the art large vision language models: Alignment,
  benchmark, evaluations and challenges.
\newblock \penalty0 (arXiv:2501.02189), 2025{\natexlab{b}}.
\newblock arXiv:2501.02189 [cs].

\bibitem[Lin et~al.(2014)Lin, Maire, Belongie, Hays, Perona, Ramanan, Dollár,
  and Zitnick]{Lin_Maire_Belongie_Hays_Perona_Ramanan_Dollar_Zitnick_2014}
Tsung-Yi Lin, Michael Maire, Serge Belongie, James Hays, Pietro Perona, Deva
  Ramanan, Piotr Dollár, and C.~Lawrence Zitnick.
\newblock Microsoft coco: Common objects in context.
\newblock In \emph{Computer Vision – ECCV 2014}, page 740–755, Cham, 2014.
  Springer International Publishing.

\bibitem[Liu et~al.(2017)Liu, Wang, and Yang]{Liu_Wang_Yang_2017}
Jingyu Liu, Liang Wang, and Ming~Hsuan Yang.
\newblock Referring expression generation and comprehension via attributes:
  16th ieee international conference on computer vision, iccv 2017.
\newblock page 4866–4874, 2017.

\bibitem[Liu et~al.(2025)Liu, Zeng, Ren, Li, Zhang, Yang, Jiang, Li, Yang, Su,
  Zhu, and Zhang]{Liu_Zeng_Ren_Li_Zhang_Yang_Jiang_Li_Yang_Su_etal._2025}
Shilong Liu, Zhaoyang Zeng, Tianhe Ren, Feng Li, Hao Zhang, Jie Yang, Qing
  Jiang, Chunyuan Li, Jianwei Yang, Hang Su, Jun Zhu, and Lei Zhang.
\newblock Grounding dino: Marrying dino with grounded pre-training for open-set
  object detection.
\newblock In \emph{Computer Vision – ECCV 2024}, page 38–55, Cham, 2025.
  Springer Nature Switzerland.

\bibitem[Mao et~al.(2016)Mao, Huang, Toshev, Camburu, Yuille, and
  Murphy]{Mao_Huang_Toshev_Camburu_Yuille_Murphy_2016}
Junhua Mao, Jonathan Huang, Alexander Toshev, Oana Camburu, Alan Yuille, and
  Kevin Murphy.
\newblock Generation and comprehension of unambiguous object descriptions.
\newblock In \emph{2016 IEEE Conference on Computer Vision and Pattern
  Recognition (CVPR)}, page 11–20, 2016.

\bibitem[Miller(1992)]{Miller_1992}
George~A. Miller.
\newblock Wordnet: A lexical database for english.
\newblock In \emph{Speech and Natural Language: Proceedings of a Workshop Held
  at Harriman, New York, February 23-26, 1992}, 1992.

\bibitem[Mimno and Thompson(2017)]{Mimno_Thompson_2017}
David Mimno and Laure Thompson.
\newblock The strange geometry of skip-gram with negative sampling.
\newblock In \emph{Proceedings of the 2017 Conference on Empirical Methods in
  Natural Language Processing}, page 2873–2878, Copenhagen, Denmark, 2017.
  Association for Computational Linguistics.

\bibitem[Min et~al.(2023)Min, Ross, Sulem, Veyseh, Nguyen, Sainz, Agirre,
  Heintz, and Roth]{Min_Ross_Sulem_Veyseh_Nguyen_Sainz_Agirre_Heintz_Roth_2023}
Bonan Min, Hayley Ross, Elior Sulem, Amir Pouran~Ben Veyseh, Thien~Huu Nguyen,
  Oscar Sainz, Eneko Agirre, Ilana Heintz, and Dan Roth.
\newblock Recent advances in natural language processing via large pre-trained
  language models: A survey.
\newblock \emph{ACM Comput. Surv.}, 56\penalty0 (2):\penalty0 30:1--30:40,
  2023.

\bibitem[Montani et~al.(2023)Montani, Honnibal, Honnibal, Boyd, Landeghem, and
  Peters]{ines_montani_2023_10009823}
Ines Montani, Matthew Honnibal, Matthew Honnibal, Adriane Boyd, Sofie~Van
  Landeghem, and Henning Peters.
\newblock explosion/spacy: v3.7.2: Fixes for apis and requirements, 2023.

\bibitem[Oquab et~al.(2024)Oquab, Darcet, Moutakanni, Vo, Szafraniec, Khalidov,
  Fernandez, HAZIZA, Massa, El-Nouby, Assran, Ballas, Galuba, Howes, Huang, Li,
  Misra, Rabbat, Sharma, Synnaeve, Xu, Jegou, Mairal, Labatut, Joulin, and
  Bojanowski]{oquab2024dinov}
Maxime Oquab, Timoth{\'e}e Darcet, Th{\'e}o Moutakanni, Huy~V. Vo, Marc
  Szafraniec, Vasil Khalidov, Pierre Fernandez, Daniel HAZIZA, Francisco Massa,
  Alaaeldin El-Nouby, Mido Assran, Nicolas Ballas, Wojciech Galuba, Russell
  Howes, Po-Yao Huang, Shang-Wen Li, Ishan Misra, Michael Rabbat, Vasu Sharma,
  Gabriel Synnaeve, Hu Xu, Herve Jegou, Julien Mairal, Patrick Labatut, Armand
  Joulin, and Piotr Bojanowski.
\newblock {DINO}v2: Learning robust visual features without supervision.
\newblock \emph{Transactions on Machine Learning Research}, 2024.
\newblock Featured Certification.

\bibitem[Peng et~al.(2023)Peng, Wang, Dong, Hao, Huang, Ma, and
  Wei]{Peng_Wang_Dong_Hao_Huang_Ma_Wei_2023}
Zhiliang Peng, Wenhui Wang, Li Dong, Yaru Hao, Shaohan Huang, Shuming Ma, and
  Furu Wei.
\newblock Kosmos-2: Grounding multimodal large language models to the world.
\newblock \penalty0 (arXiv:2306.14824), 2023.
\newblock arXiv:2306.14824 [cs].

\bibitem[Plummer et~al.(2015)Plummer, Wang, Cervantes, Caicedo, Hockenmaier,
  and Lazebnik]{Plummer_Wang_Cervantes_Caicedo_Hockenmaier_Lazebnik_2015}
Bryan~A. Plummer, Liwei Wang, Chris~M. Cervantes, Juan~C. Caicedo, Julia
  Hockenmaier, and Svetlana Lazebnik.
\newblock Flickr30k entities: Collecting region-to-phrase correspondences for
  richer image-to-sentence models.
\newblock In \emph{2015 IEEE International Conference on Computer Vision
  (ICCV)}, page 2641–2649, 2015.

\bibitem[Radford et~al.(2021)Radford, Kim, Hallacy, Ramesh, Goh, Agarwal,
  Sastry, Askell, Mishkin, Clark, Krueger, and
  Sutskever]{Radford_Kim_Hallacy_Ramesh_Goh_Agarwal_Sastry_Askell_Mishkin_Clark_2021}
Alec Radford, Jong~Wook Kim, Chris Hallacy, Aditya Ramesh, Gabriel Goh,
  Sandhini Agarwal, Girish Sastry, Amanda Askell, Pamela Mishkin, Jack Clark,
  Gretchen Krueger, and Ilya Sutskever.
\newblock Learning transferable visual models from natural language
  supervision.
\newblock In \emph{Proceedings of the 38th International Conference on Machine
  Learning}, page 8748–8763. PMLR, 2021.

\bibitem[Ren et~al.(2016)Ren, He, Girshick, and Sun]{Ren_He_Girshick_Sun_2016}
Shaoqing Ren, Kaiming He, Ross Girshick, and Jian Sun.
\newblock Faster r-cnn: Towards real-time object detection with region proposal
  networks.
\newblock \penalty0 (arXiv:1506.01497), 2016.
\newblock arXiv:1506.01497 [cs].

\bibitem[Shi et~al.(2025)Shi, Liu, Hu, Hu, Yin, and
  Hong]{Shi_Liu_Hu_Hu_Yin_Hong_2025}
Liangtao Shi, Ting Liu, Xiantao Hu, Yue Hu, Quanjun Yin, and Richang Hong.
\newblock Swimvg: Step-wise multimodal fusion and adaption for visual
  grounding.
\newblock \emph{IEEE Transactions on Multimedia}, page 1–12, 2025.

\bibitem[Vaswani et~al.(2023)Vaswani, Shazeer, Parmar, Uszkoreit, Jones, Gomez,
  Kaiser, and
  Polosukhin]{Vaswani_Shazeer_Parmar_Uszkoreit_Jones_Gomez_Kaiser_Polosukhin_2023}
Ashish Vaswani, Noam Shazeer, Niki Parmar, Jakob Uszkoreit, Llion Jones,
  Aidan~N. Gomez, Lukasz Kaiser, and Illia Polosukhin.
\newblock Attention is all you need.
\newblock \penalty0 (arXiv:1706.03762), 2023.
\newblock arXiv:1706.03762 [cs].

\bibitem[Wang et~al.(2024)Wang, He, Xuan, Sebastian, Ono, Li, Behpour, Doan,
  Gou, Shen, and
  Ren]{Wang_He_Xuan_Sebastian_Ono_Li_Behpour_Doan_Gou_Shen_etal._2024}
Xiaoqi Wang, Wenbin He, Xiwei Xuan, Clint Sebastian, Jorge~Piazentin Ono, Xin
  Li, Sima Behpour, Thang Doan, Liang Gou, Han-Wei Shen, and Liu Ren.
\newblock Use: Universal segment embeddings for open-vocabulary image
  segmentation.
\newblock In \emph{2024 IEEE/CVF Conference on Computer Vision and Pattern
  Recognition (CVPR)}, page 4187–4196, 2024.
\newblock ISSN: 2575-7075.

\bibitem[Wolfe and Caliskan(2022)]{Wolfe_Caliskan_2022}
Robert Wolfe and Aylin Caliskan.
\newblock Contrastive visual semantic pretraining magnifies the semantics of
  natural language representations.
\newblock In \emph{Proceedings of the 60th Annual Meeting of the Association
  for Computational Linguistics (Volume 1: Long Papers)}, page 3050–3061,
  Dublin, Ireland, 2022. Association for Computational Linguistics.

\bibitem[Xiao et~al.(2024)Xiao, Wu, Xu, Dai, Hu, Lu, Zeng, Liu, and
  Yuan]{Xiao_Wu_Xu_Dai_Hu_Lu_Zeng_Liu_Yuan_2024}
Bin Xiao, Haiping Wu, Weijian Xu, Xiyang Dai, Houdong Hu, Yumao Lu, Michael
  Zeng, Ce Liu, and Lu Yuan.
\newblock Florence-2: Advancing a unified representation for a variety of
  vision tasks.
\newblock In \emph{2024 IEEE/CVF Conference on Computer Vision and Pattern
  Recognition (CVPR)}, page 4818–4829, 2024.
\newblock ISSN: 2575-7075.

\bibitem[Xuan et~al.(2025)Xuan, Deng, and Ma]{Xuan_Deng_Ma_2025}
Xiwei Xuan, Ziquan Deng, and Kwan-Liu Ma.
\newblock Reme: A data-centric framework for training-free open-vocabulary
  segmentation.
\newblock \penalty0 (arXiv:2506.21233), 2025.
\newblock arXiv:2506.21233 [cs].

\bibitem[Ye et~al.(2022)Ye, Tian, Yan, Yang, Wang, Zhang, He, and
  Lin]{Ye_Tian_Yan_Yang_Wang_Zhang_He_Lin_2022}
Jiabo Ye, Junfeng Tian, Ming Yan, Xiaoshan Yang, Xuwu Wang, Ji Zhang, Liang He,
  and Xin Lin.
\newblock Shifting more attention to visual backbone: Query-modulated
  refinement networks for end-to-end visual grounding.
\newblock In \emph{2022 IEEE/CVF Conference on Computer Vision and Pattern
  Recognition (CVPR)}, page 15481–15491, 2022.

\bibitem[Young et~al.(2014)Young, Lai, Hodosh, and
  Hockenmaier]{Young_Lai_Hodosh_Hockenmaier_2014}
Peter Young, Alice Lai, Micah Hodosh, and Julia Hockenmaier.
\newblock From image descriptions to visual denotations: New similarity metrics
  for semantic inference over event descriptions.
\newblock \emph{Transactions of the Association for Computational Linguistics},
  2:\penalty0 67–78, 2014.

\bibitem[Yu et~al.(2016)Yu, Poirson, Yang, Berg, and
  Berg]{Yu_Poirson_Yang_Berg_Berg_2016}
Licheng Yu, Patrick Poirson, Shan Yang, Alexander~C. Berg, and Tamara~L. Berg.
\newblock Modeling context in referring expressions.
\newblock In \emph{Computer Vision – ECCV 2016}, page 69–85, Cham, 2016.
  Springer International Publishing.

\bibitem[Yu et~al.(2018)Yu, Lin, Shen, Yang, Lu, Bansal, and
  Berg]{Yu_Lin_Shen_Yang_Lu_Bansal_Berg_2018}
Licheng Yu, Zhe Lin, Xiaohui Shen, Jimei Yang, Xin Lu, Mohit Bansal, and
  Tamara~L. Berg.
\newblock Mattnet: Modular attention network for referring expression
  comprehension.
\newblock In \emph{2018 IEEE/CVF Conference on Computer Vision and Pattern
  Recognition}, page 1307–1315, 2018.
\newblock ISSN: 2575-7075.

\bibitem[Yu and Li(2024)]{Yu_Li_2024}
Zhihan Yu and Ruifan Li.
\newblock Revisiting counterfactual problems in referring expression
  comprehension.
\newblock In \emph{2024 IEEE/CVF Conference on Computer Vision and Pattern
  Recognition (CVPR)}, page 13438–13448, 2024.
\newblock ISSN: 2575-7075.

\end{thebibliography}
}

\end{document}